\documentclass[acmtog, authorversion]{acmartOURS}

\setcitestyle{square}

\acmJournal{TOG}

\setcopyright{acmcopyright}

\acmDOI{}

\usepackage{steinmetz}
\usepackage{graphicx}
\usepackage{amssymb}
\usepackage{amsmath}
\usepackage{mathtools}
\usepackage{multirow}
\usepackage{algorithm}
\usepackage{algorithmic}
\usepackage{newlfont} 
\usepackage{fixltx2e}
\usepackage{subfig} 
\usepackage[english]{babel} 
\usepackage{enumitem}
\usepackage{microtype}
\usepackage{floatflt}
\usepackage{booktabs} 

\makeatletter
\newcommand\footnoteref[1]{\protected@xdef\@thefnmark{\ref{#1}}\@footnotemark}
\makeatother

\newcommand{\nothing}[1]{}

\newcommand{\rev}[1]{{#1}}
\newcommand{\revB}[1]{{#1}}

\definecolor{HaibinColor}{rgb}{0.7,0,0} 

\definecolor{ErsinColor}{rgb}{0,0.6,0} 

\definecolor{VangelisColor}{rgb}{0,0,0.8} 

\definecolor{DuyguColor}{rgb}{0.9,0.44,0.0} 

\definecolor{VovaColor}{rgb}{0.56,0.34,0.62} 

\definecolor{SidColor}{rgb}{1.0,0.3,0.3} 

\definecolor{ReviewerColor}{rgb}{0.3,0.3,1.0} 

\newcommand{\pseudocode}{Program}
\floatstyle{plain}
\newfloat{algorithm}{tbhp}{lop}
\floatname{algorithm}{\pseudocode}

\ifdefined\mainonly
\else
\newcommand{\mainonly}{0}
\fi

\begin{document}

\title{Learning local shape descriptors from part correspondences with multi-view convolutional networks}
\author{Haibin Huang}
\affiliation{ \institution{University of Massachusetts Amherst} }
\author{Evangelos Kalogerakis}
\affiliation{ \institution{University of Massachusetts Amherst} }
\author{Siddhartha Chaudhuri}
\affiliation{ \institution{IIT Bombay} }
\author{Duygu Ceylan}
\affiliation{ \institution{Adobe Research} }
\author{Vladimir G. Kim}
\affiliation{ \institution{Adobe Research} }
\author{Ersin Yumer}
\affiliation{ \institution{Adobe Research} }

\renewcommand\shortauthors{Huang, H. et al}


\begin{abstract}
We present a new local descriptor for 3D shapes, directly applicable to a wide range of shape analysis problems such as point correspondences, semantic segmentation, affordance prediction, and shape-to-scan matching.  \rev{The descriptor is produced by a convolutional network that is trained to embed geometrically and semantically similar points close to one another in descriptor space. The network processes surface neighborhoods around points on a shape that are captured at multiple scales by a succession of progressively zoomed out views, taken from carefully selected camera positions. We leverage two extremely large sources of data to train our network. First, since our network processes rendered views in the form of 2D images, we repurpose architectures pre-trained on massive image datasets. \revB{Second, we automatically generate a synthetic dense point correspondence dataset by non-rigid alignment of  corresponding shape parts in a large collection of segmented 3D models}. As a result of these design choices, our network effectively encodes multi-scale local context and fine-grained surface detail. \revB{Our network can be  trained to produce either category-specific descriptors or more generic descriptors by learning from  multiple shape categories. Once trained, at test time, the network  extracts local descriptors for shapes without requiring any part segmentation as input. Our method can  produce effective local descriptors  even for shapes whose category is unknown or different from the ones used while training.}
We demonstrate through several experiments that our learned local descriptors are more discriminative compared to state of the art alternatives, and are effective in a variety of shape analysis applications.}

\end{abstract}

%
%


%
%


\begin{teaserfigure}
  \includegraphics[width=\textwidth]{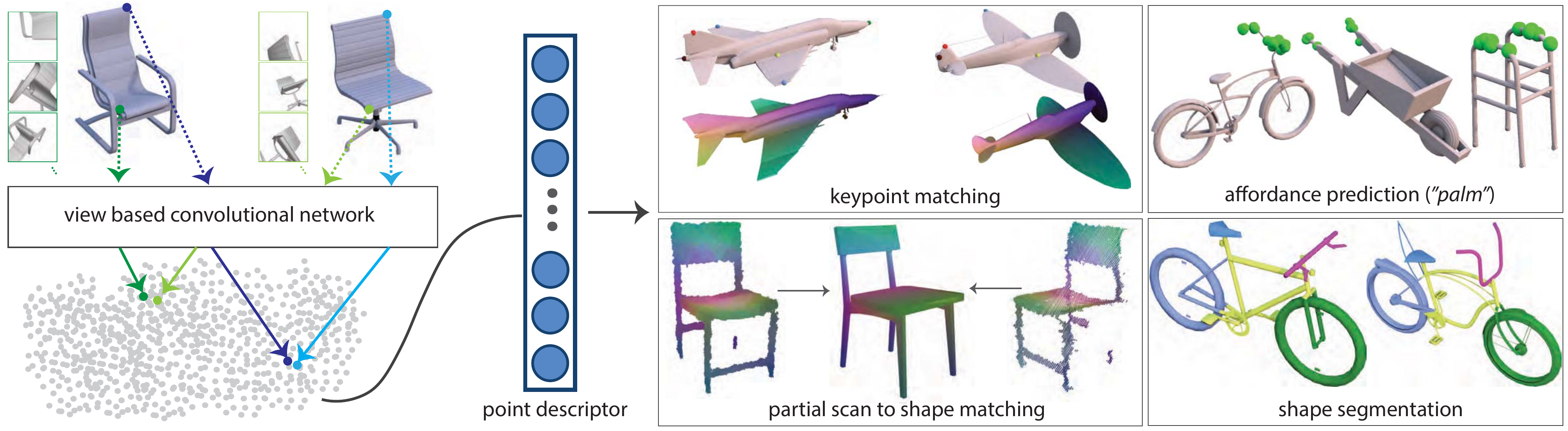}
  \caption{We present a view-based convolutional network that produces local, point-based shape descriptors. The network is trained such that geometrically and semantically similar points across different 3D\ shapes are embedded close to each other in descriptor space (left). Our produced descriptors are quite generic --- they can be used in a variety of shape analysis applications, including dense matching, prediction of human affordance regions, partial scan-to-shape matching, and shape segmentation (right).}
  \label{fig:teaser}
\end{teaserfigure}

\maketitle


\section{Introduction}

Local descriptors for surface points are at the heart of a huge variety of 3D shape analysis problems such as keypoint detection, shape correspondence, semantic segmentation, region labeling, and 3D reconstruction ~\cite{Xu:2016}. The vast majority of algorithms addressing these problems are predicated on local surface descriptors. Such descriptors characterize the local geometry around the point of interest in a way that the geometric similarity between two points can be estimated by some simple, usually Euclidean, metric in descriptor space. A large number of descriptors have been developed for specific scenarios  which can encode both local analytical properties like curvature as well as some notion of the larger context of the point within the shape, such as the relative arrangement of other shape regions.

Yet the aim in shape analysis is frequently not geometric but functional,  or ``semantic'', similarity of points and local shape regions. Most existing descriptors rely on two main assumptions: (a)~one can directly specify which geometric features are relevant for semantic similarity, and (b)~strong post-process regularization, such as dense surface alignment, can compensate for all situations when the first assumption fails~\cite{Xu:2016}. The latter is typically computationally expensive, difficult to develop, and task-specific: it benefits greatly from access to better descriptors that reduce the post-processing burden. In this work, we challenge the first assumption. \rev{As several examples in this paper show, it is hard to hand-craft descriptors that effectively capture functional, or ``semantic'' shape properties since these  are often a complex function of local geometry, structure and context - for example, see the geometric and structural variability of ``palm affordance'' object regions in Figure \ref{fig:teaser} (i.e., regions where humans  tend to place their palms when they interact with these objects). An additional challenge in this functional, or ``semantic'' shape analysis is that local descriptors should be invariant to structural variations of shapes (e.g., see keypoint matching and segmentation in Figure \ref{fig:teaser}), and should be robust to missing data, outliers and noise (e.g., see partial scans in Figure \ref{fig:teaser}).}

\rev{Since it is hard to decide a-priori which aspects of shape geometry are more or less relevant for point similarity, we adopt a learning approach to automatically {\em learn} a local descriptor that implicitly captures a notion of higher level similarity between points, while remaining robust to structural variations, noise, and differences in data sources and representations. We also aim to achieve this in a data-driven manner, relying on nothing other than examples of corresponding points on pairs of different shapes. Thus, we do not need to manually guess what geometric features may be relevant for correspondence: we deduce it from examples.}

\rev{Recent works have explored the possibility of learning local descriptors that are robust to natural shape deformations~\cite{masci2015geodesic,Monti2017} or are adaptable to new data sources~\cite{zeng20163dmatch}. Yet in contrast to  findings in the image analysis community where learned descriptors are ubiquitous and general~\cite{Simo-Serra:2015,Xufeng:2015,Yi16LIFT}, learned 3D descriptors have not been as powerful as 2D counterparts because they (1)~ rely on limited training data originating from small-scale shape databases, (2)~ are computed at low spatial resolutions resulting in loss of detail sensitivity, and (3) are designed to operate on specific shape classes, such as deformable shapes.}


To overcome these challenges, we introduce a multi-scale, view-based, projective representation for local descriptor learning on 3D shapes. Given a mesh or a point cloud, our method produces a local  descriptor for any point on the shape. We represent the query point by a set of rendered views around it, inspired by the approach of Su et al.~\shortcite{SuMKL15} for global shape classification. To better capture local and global context, we render views at multiple scales and propose a novel method for viewpoint selection that avoids undesired self-occlusions. This representation naturally lends itself to 2D convolutional neural networks (CNN) operating on the views. The final layer of the base network produces a feature vector for each view, which are then combined across views via a pooling layer to yield a single descriptive vector for the point. The network is trained in a Siamese fashion~\cite{Bromley:1993} on training pairs of corresponding points.

The advantages of our approach are two-fold. {\em First}, the spatial resolution of the projected views is significantly higher than that of voxelized shape representations, which is crucial to encoding local surface details while factoring in global context. {\em Second}, 2D rendered views are similar to natural images, allowing us to repurpose neural network architectures that have achieved spectacular success in 2D computer vision tasks. We initialize our framework with filters from a base network for classifying natural images~\cite{krizhevsky2012imagenet}, whose weights are pretrained on over a million image exemplars~\cite{ILSVRC15}.

To fine-tune the network architecture for descriptor learning, we require access to training pairs of semantically similar points. Here, we make another critical observation. Although correspondence data is available only in limited quantities, large repositories of consistently segmented and labeled shapes have recently become available~\cite{Yi16}. We can rely on semantic {\em segmentation} to guide a part-aware, non-rigid alignment method for semantic {\em correspondence}, in order to generate very large amounts of dense training data. \rev{Our synthetic dataset of $\sim$977M corresponding point pairs is the largest such repository, by several orders of magnitude, assembled so far for learning.}

To summarize, the contributions of this paper are:
\begin{itemize}
\item A new point-based, local descriptor for general 3D shapes, directly applicable to a wide range of shape analysis tasks, that is sensitive to both  fine-grained local information and context.
\item A  convolutional network architecture for combining rendered  views around a surface point at multiple scales into a single compact, local descriptor.
\item A method for per-point view selection that avoids self-occlusions and provides a collection of informative rendered projections.
\item A massive synthetic dataset of corresponding point pairs for training purposes.
\end{itemize}

We demonstrate that our descriptors can be directly used in many applications, including key point detection, affordance labeling for human interaction, and shape segmentation tasks on complete 3D shapes. Further, they can be used for partial shape matching tasks on unstructured scan data {\em without any fine-tuning}. We evaluate our method on sparse correspondence and shape segmentation benchmarks and demonstrate that our point-based descriptors are significantly better than traditional hand-crafted and voxel-based shape descriptors.

\begin{figure*}[t]
    \centering
    \includegraphics[width=1.0\linewidth]{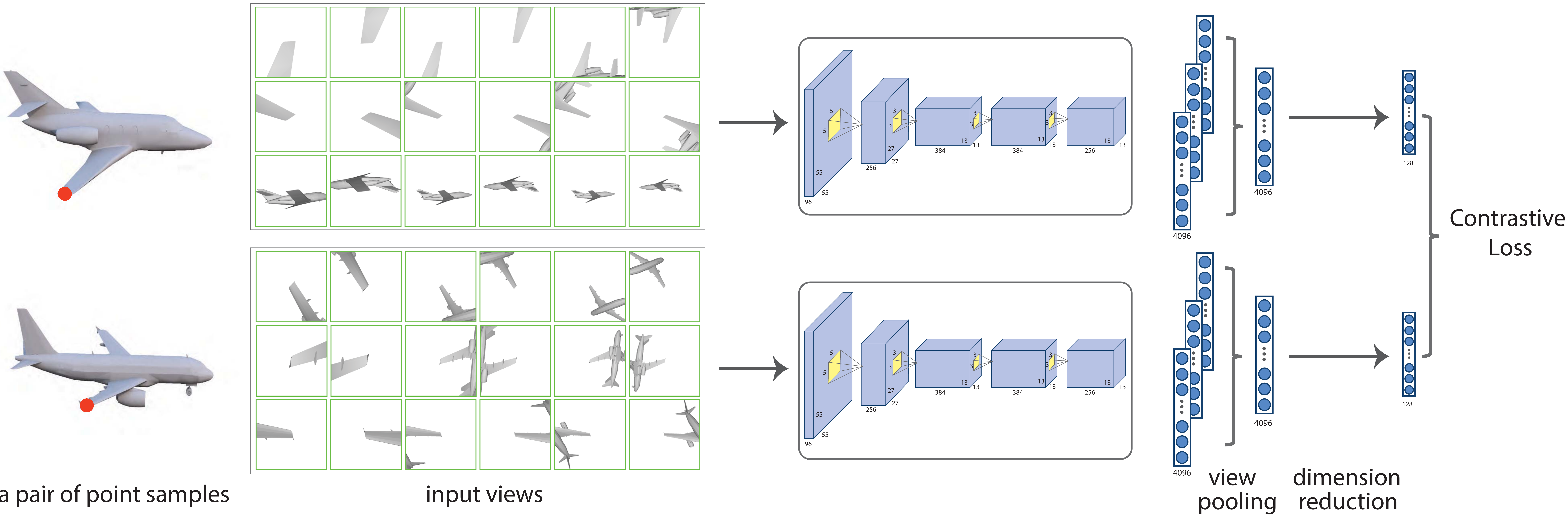}
    \caption{Given a pair of surface points (shown in red color) on two shapes, we generate a set of rendered images that capture the local context around each  point in multiple scales. Each set of  images is processed through an identical stack of convolutional, pooling and non-linear transformation layers resulting in a $4096$ dimensional descriptor per point. We aggregate these descriptors across all the input views by using a \emph{view pooling} strategy and further reduce the dimensionality of the descriptor. We train this network with an objective function (contrastive loss) to ensure that geometrically and semantically similar (or dissimilar) point pairs are assigned similar (or dissimilar) descriptors.}
    \label{fig:architecture}
\end{figure*}

\section{Related work}
Representing 3D geometric data with global or local descriptors is a longstanding problem in computer graphics and vision with several applications. Global analysis applications, such as recognition and retrieval, require global descriptors that map every \emph{shape} to a point in a descriptor space. Local analysis, such as dense feature matching or keypoint detection, needs local descriptors that map every \emph{point on a shape} to a point in descriptor space. In this section we present a brief overview of traditional hand-crafted descriptors and learned representations.

\paragraph{Traditional shape descriptors.}
\rev{Traditionally, researchers and engineers relied on their intuition to hand-craft shape descriptors. Examples of global shape descriptors include 3D shape histograms~\cite{Ankerst:1999}, spherical harmonics \cite{Saupe:2001:MRS}, shape distributions \cite{Osada:2002:SD}, light-field descriptors~\cite{Chen:2003}, 3D Zernike moments \cite{Novotni:2003:ZDC}, and symmetry-based features~\cite{Kazhdan:2004:SDA}. Examples of local per-point shape descriptors include spin images~\cite{johnson1999using}, shape contexts \cite{Belongie02}, geodesic distance functions~\cite{Zhang:2005}, curvature features \cite{Gal:2006:SGF}, histograms of surface normals \cite{Tombari:2010:USH}, shape diameter \cite{Shapira:2010:CPA}, PCA-based descriptors \cite{kalogerakis2010learning}, heat kernel descriptors~\cite{Bronstein:2011}, and wave kernel signatures \cite{Aubry2011,Rodola2014}. These descriptors  capture low-level geometric information, which often cannot be mapped reliably to functional or ``semantic'' shape properties. In addition, these descriptors often lack robustness to noise, partial data, or large structural shape variations. They frequently rely on a small set of hand-tuned parameters, which are tailored  for specific datasets or shape processing scenarios. We refer the reader to the recent survey of Xu et al.~\shortcite{Xu:2016} for a more detailed discussion on hand-crafted descriptors.}

\paragraph{Learned global descriptors.} With the recent success of learning based methods, specifically deep neural networks, there has been a growing trend to learn descriptors directly from the data itself instead of manually engineering them. This trend has become apparent in the vision community with the proposal of both global~\cite{krizhevsky2012imagenet} and local~\cite{Simo-Serra:2015,Xufeng:2015,Yi16LIFT} deep image descriptors.

\rev{With the availability of large 3D shape collections~\cite{Chang:2015}, we have seen a similar trend to learn 3D shape descriptors, mostly focusing on learning global descriptors. Early efforts involved shallow metric learning \cite{Ohbuchi:2010:DML} and dictionary learning based on clustering  (bags-of-words models) \cite{Liu:2006:ST, Lavoue:2012:CBD} or sparse coding \cite{Litman:2014:SLB}. As a direct extension of the widely successful deep learning techniques in 2D image grids to 3D, voxel-based shape representations have been widely adopted in deep learning for 3D shape recognition~\cite{Maturana_2015_7900,WuSKYZTX15,Qi:2016a,DeepSlidingShapes}. An alternative approach is to first extract a set of hand-crafted geometric features, then further process them through a deep neural network for shape classification and retrieval \cite{Xie2015}. Sinha et al.~\shortcite{Sinha2016} apply deep networks on global shape embeddings in the form of geometry images. In the context of RGB-D images, neural networks have been proposed to combine features learned from both RGB images and depth data~\cite{Socher:2012,Bo:2014,lai_icra14}. All the above-mentioned learned global 3D features have shown promising results for 3D object detection, classification, and retrieval tasks. Our focus in this paper, on the other hand, is to learn point-based shape descriptors that can be used for dense feature matching, keypoint detection, affordance labeling and other local shape analysis applications.}

\rev{Our method is particularly inspired by Su et al.'s multi-view convolutional network for shape classification \cite{SuMKL15}, which has demonstrated high performance in  recent shape classification and retrieval benchmarks \cite{Savva:2016:LSR}. Multi-view architectures have also been employed recently for shape segmentation \cite{kalogerakis2017}. These architectures are designed to produce a single representation for an entire shape \cite{SuMKL15}, or part label confidences \cite{kalogerakis2017}. In contrast, our architecture is designed to produce surface point descriptors. Instead of optimizing for a shape classification or part labeling objective, we employ a siamese architecture that compares  geometric similarity of points during training. Instead of fixed views \cite{SuMKL15}, or views selected to maximize surface coverage  \cite{kalogerakis2017}, we use local views adaptively selected to capture multi-scale context around surface points. We also propose an automatic method to create a massive set of point-wise correspondence data to train our architecture based on part-guided, non-rigid registration.}

\paragraph{Learned surface descriptors.} \rev{Recently, there have been some efforts towards learning surface point descriptors. In the context of deformable models, such as human bodies, deep learning architectures have been designed to operate on intrinsic   surface representations \cite{masci2015geodesic,Boscaini:2015,Boscaini2016,Monti2017}. The learned intrinsic descriptors produced by these architectures exhibit invariance to isometric or near-isometric deformations. However, in several shape classes, particularly man-made objects, even rigid rotations of parts can change their underlying functionality and  semantic correspondences to other parts (e.g. rotating a horizontal tailplane 90 degrees in an airplane would convert it into a vertical stabilizer). Our network attempts to learn the invariance to shape deformations if and when such invariance exists. We also note that in a  recent large-scale benchmark \cite{shrec2017},\ developed concurrently to our work, learning-based extrinsic methods  seem to  outperform learning-based intrinsic methods  in the case of deformable shapes with missing parts. In another concurrent work, Yi et al.\shortcite{Yi2017} synchronize the spectral domains of shapes to learn intrinsic descriptors that are more robust to large structural variations of shapes. To initialize this synchronization, they assume pre-existing extrinsic alignments between shapes. In our case, we do not make any assumptions about consistent shape alignment or orientation.

Wei et al.~\shortcite{wei2016dense} learn feature descriptors for each pixel in a depth scan of a human for establishing dense correspondences. Their descriptors are tuned to classify pixels into 500 distinct regions or 33 annotated keypoints on human bodies. They are extracted per pixel in a single depth map (and view). In contrast, we produce a single, compact representation of a 3D point by aggregating information across multiple views.  Our representation is tuned to compare similarity of 3D points between shapes of different structure or even functionality, going well beyond human body region classification. The method of Guo et al.~\shortcite{Guo:2015} first extracts a set of hand-crafted geometric descriptors, then utilizes neural networks to map them to part labels. Thus, this method still inherits the main limitations of hand-crafted descriptors. In a concurrent work, Qi et al.~\shortcite{Qi:2017} presented a classification-based network architecture that directly receives an unordered set of input point positions in 3D and learns shape and part labels. However, this method relies on augmenting local per-point descriptors with global shape descriptors, making it more sensitive to global shape input. Zeng et al.~\shortcite{zeng20163dmatch}  learns a local volumetric patch descriptor for RGB-D data. While they show impressive results for aligning depth data for reconstruction, limited training data and limited resolution of voxel-based surface neighborhoods still remain key challenges in this approach. Instead, our deep network architecture operates on view-based projections of local surface neighborhoods at multiple scales, and adopts image-based processing layers pre-trained on massive image datasets. We refer to the evaluation section (Section~\ref{sec:evl}) for a direct comparison with this approach.}

\rev{Also related to our work is the approach by Simo-Serra et al. \cite{Simo-Serra:2015}, which learns representations for \mbox{$64 \times 64$} natural image patches through a Siamese architecture, such that patches depicting the same underlying 3D surface point tend to have similar (not necessarily same) representation across different viewpoints. In contrast, our method aims to learn surface descriptors such that geometrically and semantically similar points across different shapes are assigned similar descriptors. Our method learns a single, compact representation for a 3D surface point (instead of an image patch) by explicitly aggregating information from multiple views and at multiple scales through a view-pooling layer in a much deeper network. Surface points can be directly compared through their learned descriptors, while Simo-Serra et al. would require comparing image descriptors for all pairs of views between two 3D points, which would be computationally very expensive.}

\section{Overview}

The goal of our method is to provide a function $f$ that takes as input any surface point $p$ of a 3D shape and outputs a descriptor $X_p \in \mathbb{R}^D$ for that point, where $D$ is the output descriptor dimensionality. The function is designed such that descriptors of geometrically and semantically similar surface points across shapes with different structure are as close as possible to each other (under the Euclidean metric). Furthermore, we favor rotational invariance of the function   i.e. we do not restrict our input shapes to have consistent alignment or any particular orientation. Our main assumption is that the input shapes are represented as polygon meshes or point clouds without any restrictions on their connectivity or topology.

We  follow a machine learning approach to automatically infer this function from training data. \rev{The function can be learned either as a category-specific one (e.g. tuned for matching points on chairs) or as a cross-category one (e.g. tuned to match human region affordances across chairs, bikes, carts and so on).} At the heart of our method lies a convolutional neural network (CNN) that aims to encode this function through multiple, hierarchical processing stages involving convolutions and non-linear transformations.
\paragraph{View-based architecture.} The architecture of our CNN is depicted in Figure~\ref{fig:architecture} and described in detail in Section~\ref{sec:arch}. Our network takes as input \rev{an unordered set of} 2D\ perspective projections (rendered views) of surface neighborhoods capturing local context around each surface point in multiple scales. At a first glance, such input representation might appear non-ideal due to potential occlusions and lack of desired surface parametrization properties, such as isometry and bijectivity. On the other hand, this view-based representation is closer to human perception (humans perceive projections of 3D shape surfaces), and allows us to directly re-purpose image-based CNN architectures trained on massive image datasets. Since images depict shapes of photographed objects (along with texture), convolutional filters in these image-based architectures already partially encode shape information. Thus, we initialize our architecture using these filters, and further fine-tune them for our task. This initialization strategy has provided superior performance in other 3D shape processing tasks, such as shape classification and retrieval \cite{SuMKL15}. In addition, to combat surface information loss in the input view-based shape representation, our architecture takes as input multiple, local perspective projections per surface point, carefully chosen so that each point is always visible in the corresponding views. Figure~\ref{fig:architecture} shows the images used as input to our network for producing a descriptor on 3D airplane wingtip points.

\paragraph{Learning.} Our method automatically learns the network parameters based on a training dataset. To ensure that the function encoded in our network is general enough, a large corpus of automatically generated shape training data is used, as described in Section~\ref{sec:learning}. Specifically,  the parameters are learned such that the network favors two properties: (i) pairs of semantically similar points 
are embedded nearby in the descriptor space, and (ii) pairs of semantically dissimilar points
are separated  by a minimal constant margin in the descriptor space.
To achieve this, during the learning stage, we sample pairs of surface points from our training dataset, and process their view-based representation through two identical, or ``Siamese'' branches of our network to output their descriptors and measure their distance (see Figure~\ref{fig:architecture}).

\paragraph{Applications.} We demonstrate the effectiveness of the local descriptors learned by our architecture on a variety of geometry processing applications including labeling shape parts, finding human-centric affordances across shapes of different categories, and matching shapes to depth data (see Section~\ref{sec:app}). 

\section{Architecture}
\label{sec:arch}
We now describe our pipeline and network architecture (Figure~\ref{fig:architecture}) for extracting a local descriptor per surface point. To train the network in our implementation, we uniformly sample the input shape surface  with $1024$ surface points, and compute a descriptor for each of these points. We note that during test time we can sample any arbitrary point on a shape and compute its descriptor.

\begin{figure}[t]
    \centering
    \includegraphics[width=1.0\linewidth]{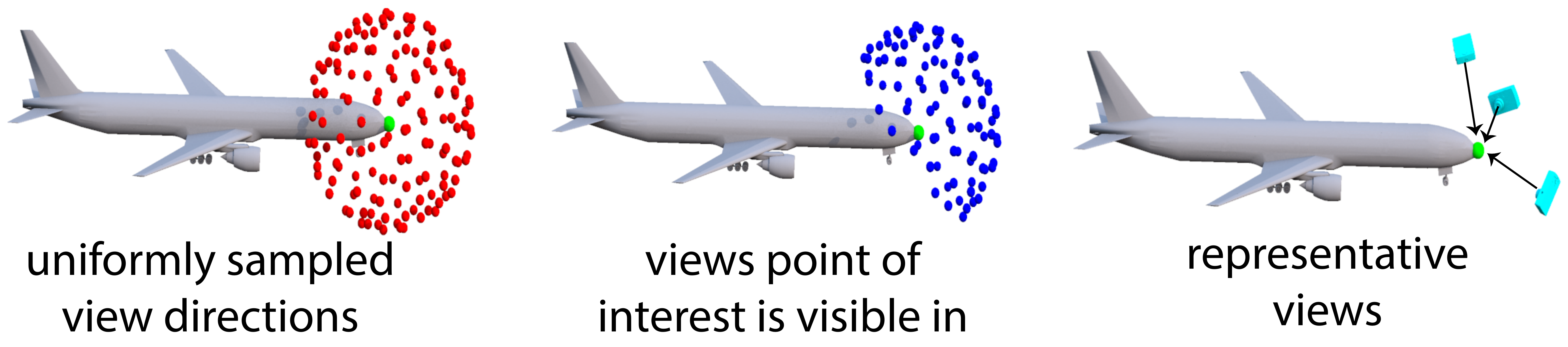}
    \caption{For a given point (in green), we show uniformly sampled viewpoints around the shape (in red). We identify the subset of these viewing directions the point is visible from (in blue). Then we perform view clustering to select 3 representative viewing directions. }
    \label{fig:view_selection}
\end{figure}

\paragraph{Pre-processing.} In the pre-processing stage, we first uniformly sample viewing directions around the shape parameterized by spherical coordinates ($\theta$, $\phi$) ($150$ directions in our implementation). We render the shape from each viewing direction such that each pixel stores indices to surface points mapped onto that pixel through perspective projection. As a result, for each surface sample point, we can  retrieve the viewing directions from which the point is visible. Since neighboring viewing directions yield very similar rendered views of surface neighborhoods, we further prune the set of viewing directions per surface sample point, significantly reducing the number of redundant images fed as input into our network. Pruning is done by executing the K-mediods clustering algorithm on the selected viewing directions  (we use their spherical coordinates to measure spherical distances between them). We set $K=3$ to select representative viewing directions  (Figure~\ref{fig:view_selection}).
To capture multi-scale contexts for each surface point and its selected viewing directions, we create $M=3$ viewpoints placed at distances $0.25$, $0.5$, $0.75$ of the shape's bounding sphere radius.  
We experimented with various viewpoint distance configurations. The above configuration yielded the most robust descriptors, as discussed in Section \ref{sec:evl}. Increasing the number of viewing directions $K$ or number of distances $M$ did not offer significant improvements.
\paragraph{Input.} The input to our network is a set of rendered images depicting local surface neighborhoods around a surface point $p$ based on the viewpoint configuration described above. Specifically, we render the shape surface around $p$ from each of the selected viewpoints, using a Phong shader and a single directional light (light direction is set to viewpoint direction). Since rotating the 3D shape would result in rotated input images, to promote rotational invariance, we rotate the input images $L=4$ times at 90 degree intervals (i.e, 4 in-plane rotations), yielding in total $K \times M \times L = 36$ input images per point. Images are rendered at $227 \times 227$ resolution.

\paragraph{View-pooling.} Each of the above input images is processed through an identical stack of convolutional, pooling and non-linear transformation layers. Specifically, in our implementation, our CNN follows the architecture known as AlexNet~\cite{krizhevsky2012imagenet}. It includes two convolutional layers, interchanged with two pooling layers and ReLu nonlinearities, followed by three additional convolutional layers with ReLu nonlinearities, a pooling layer and a final fully connected layer. We exclude the last two fully connected layers of Alexnet, (``fc7''), (``fc8''). The last layer is related to image classification on ImageNet, while using the penultimate layer did not improve the performance of our method as shown in our experiments (see Section \ref{sec:evl}).


Passing each rendered image through the above architecture yields a $4096$ dimensional descriptor. Since we have 36 rendered views in total per point, we need to aggregate these 36 image-based descriptors into a single, compact point descriptor. The reason is that evaluating the distance of every single image-based descriptor of a point with all other 36 image-based descriptors of another point (1296 pairwise comparisons) would be prohibitively expensive. Note that these 36 views are not ordered in any manner, thus there is no one-to-one correspondence between views and the image-based descriptors of different points (as discussed earlier, shapes are not consistently aligned, points are unordered, thus viewing directions are not consistent across different points).

To produce a single point descriptor, we aggregate the descriptors across the input $36$ rendered views, by using an element-wise maximum operation that selects the most discriminative descriptors (largest feature entries) across views. A similar strategy of ``max-view pooling'' has also been effectively used for shape recognition \cite{SuMKL15} - in our case, the pooling is applied to local (rather than global) image-based descriptors. Mathematically, given 36 image-based descriptors $Y_{v,p} \in \mathbb{R}^{4096}$ of a point $p$ for views $v=1 \dots 36$, max view-pooling yields a single descriptor $Y_{p} \in \mathbb{R}^{4096}$ as follows: $ Y_{p} = \underset{v}{max}(Y_{v,p} ) $.

\rev{We also experimented with taking the average image-based descriptor values across views (``average'' view-pooling) as a baseline. However, compared to ``max'' view-pooling, this strategy led to worse correspondence accuracy, as discussed in Section \ref{sec:results}. An alternative strategy would be to concatenate all the view descriptors, however, this would depend on an  ordering of all views. Ordering the views would require consistent local coordinate frames for all surface points, which is not trivial to achieve.}

\paragraph{Dimensionality reduction.} Given the point descriptor $Y_p$ produced by view-pooling aggregation, we further reduce its dimensionality to make nearest neighbor queries more efficient and also down-weigh any dimensions that contain no useful information (e.g. shading information). Dimensionality reduction is performed by adding one more layer in our network after view pooling that performs a linear transformation: $X_p = W \cdot Y_p$, where $W$ is a learned matrix of size $K \times 4096$, where $K$ is the  dimensionality of the output descriptor. The output dimensionality $K $ was selected by searching over a range of values $K=16, 32, 64, ..., 512$ and examining performance in a hold-out validation dataset. Based on our experiments, we selected $K= 128$ (see Section \ref{sec:results}).

\section{Learning}
\label{sec:learning}
Our learning procedure aims to automatically estimate the parameters of the function encoded in our deep architecture. The key idea of our learning procedure is to train the architecture such that it produces similar descriptor values for points that are deemed similar in geometric and semantic sense. To this end, we require training data composed of corresponding pairs of points. One possibility is to define  training correspondences by hand, or resort to crowd-sourcing techniques to gather such correspondences. Our function has millions of parameters (40M), thus gathering a large enough dataset with millions of correspondences, would require a significantly large amount of human labor, plus additional human supervision to resolve conflicting correspondences. Existing correspondence benchmarks~\cite{kim2013learning} have limited number of shapes, or focus on specific cases, e.g. deformable shapes~\cite{Bogo:CVPR:2014}.

We instead generate training correspondences automatically by leveraging highly structured databases of consistently segmented shapes with labeled parts. The largest such database is the segmented ShapeNetCore dataset~\cite{Yi16} that includes 17K man-made shapes distributed in 16 categories. Our main observation is that while these man-made shapes have significant differences in the number and arrangement of their parts, individual parts with the same label are often related by simple deformations~\cite{Ovsjanikov:2011:ECV}. By computing these deformations through non-rigid registration executed on pairs of parts with the same label, we can get a large dataset of training point correspondences. \rev{Even if the resulting correspondences are potentially not as accurate as carefully human-annotated point correspondences, their massive quantity tends to counterbalance any noise and imperfections.} In the next paragraphs, we discuss the generation of our training dataset of correspondences and how these are used to train our architecture.

\begin{figure}[t]
    \centering
    \includegraphics[width=\columnwidth]{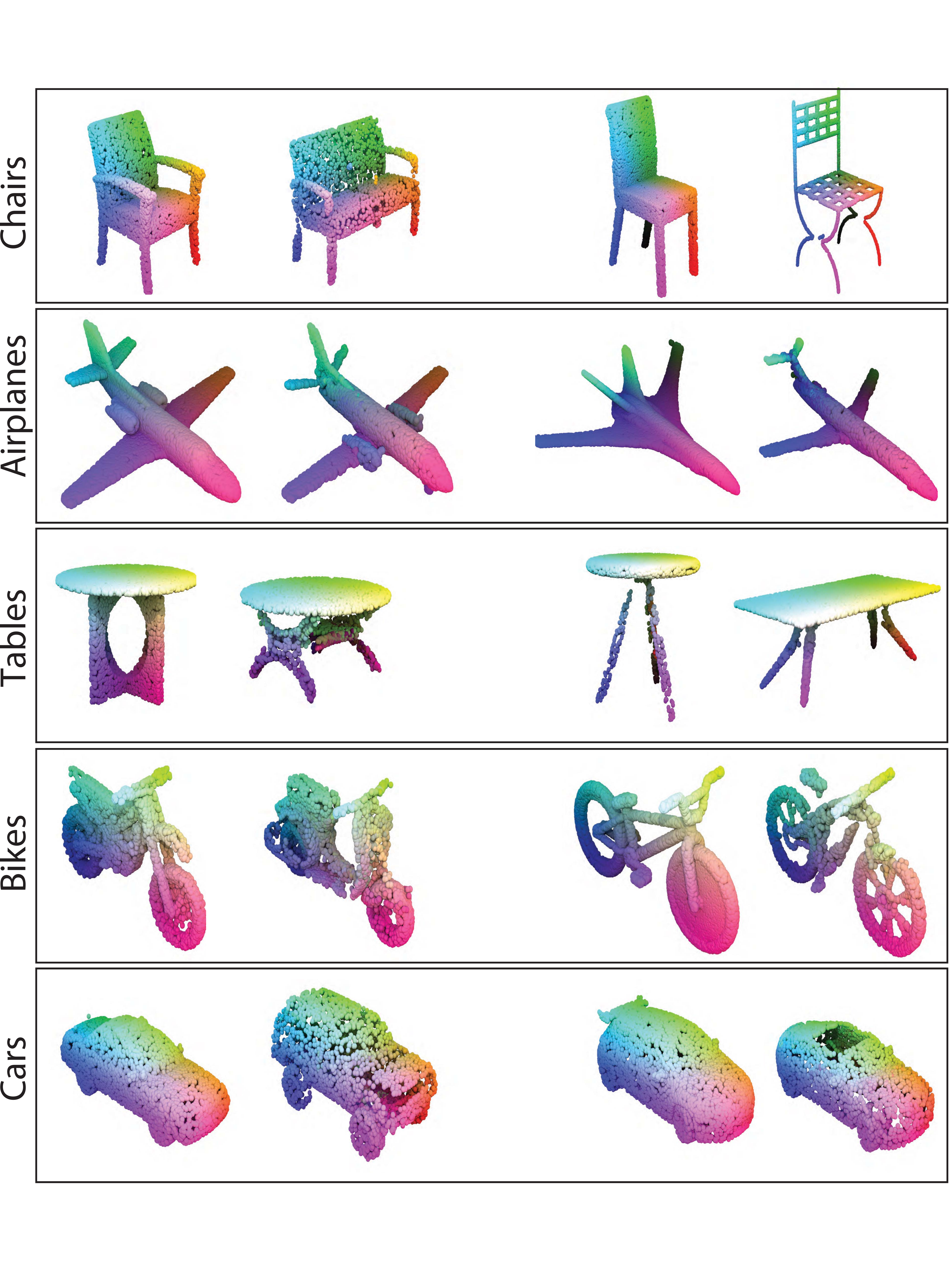}
    \vspace{-7mm}
    \caption{Visualization of training point correspondences computed through part-based registration for pairs of shapes from ShapeNetCore. Corresponding points are shown in similar colors.}
    \label{fig:syntheticData}
    \vspace{-3mm}    
\end{figure}

\paragraph{Part-based registration.}


Given a pair of consistently segmented shapes $A$ and $B$, our registration algorithm aims to non-rigidly align all pairs of segments with the same label in the two shapes. First, we sample $10K$
points $P_A$ and $P_B$ from their mesh representation. The points are tagged with the part labels of their corresponding faces. Let $P^\ell_A, P^\ell_B$ denote the sets of points originating from a pair of corresponding parts with the same label $\ell$ in $A$ and $B$ respectively. For each part, we compute an initial affine transformation $T_\ell$ for $P^\ell_A$ so that it has the same oriented bounding box as $P^\ell_B$. Then for each point $a \in P^\ell_A$, we seek a translation (offset) $o(a)$ that moves it as-close-as possible to the surface represented by $P^\ell_B$, while offsets for neighboring points in $P^\ell_A$ are as similar as possible to ensure a smooth deformation. In the same manner, for each point $b \in P^\ell_B$, we compute an offset $o(b)$ that smoothly deforms the part $P^\ell_B$ towards  $P^\ell_A$. To compute the offsets, we minimize a deformation energy that penalizes distances between the point sets of the two parts, and inconsistent offsets between neighboring points:
\begin{align}
E_{def} &=
\sum\limits_{a \in P^\ell_A} dist^{2}(a+o(a), P^\ell_B)+\sum\limits_{b \in P^\ell_B} dist^{2}(b+o(b), P^\ell_A) \nonumber \\
&+ \sum\limits_{a,a' \in N(a)} ||o(a) -o(a')||^2 + \sum\limits_{b,b' \in N(b)} ||o(b) -o(b')||^2
\end{align}
where $N(a),N(b)$ are neighborhoods for each point $a,b$ respectively (in our implementation, we use 6 nearest neighbors per point), and $dist$ computes the distance of a translated point to the closest compatible point of the other point set. The energy can be minimized using an ICP-based procedure: given closest pairs of compatible points on the two parts initially, offsets are computed by minimizing the above energy, then closest pairs are updated. The final offsets provide a dense correspondence between closest compatible points of $A$ and $B$.

Altough alternative deformation procedures could be used (e.g. as-rigid-as possible deformations
~\cite{Sorkine:2007:ASM,Sumner:2007:EDS}), we found that our technique provides satisfactory pairs (Figure~\ref{fig:syntheticData}), and  is fast enough to provide a massive dataset: $100K$ pairs of parts with $10K$ points each were aligned in $12$ hours on 3 CPUs with $14$ hyperthreaded cores each. Table \ref{tab:train_stats} lists statistics about the training dataset we created for 16 ShapeNetCore classes.

\begin{table}[t]
\centering
\begin{tabular}{c|c|c|c}
\multirow{1}{*}{ShapeNetCore} & \multirow{1}{*}{\# shapes} & \multirow{1}{*}{\# aligned} & \multirow{1}{*}{\# corresponding}\\
    Category                      & used                    &  shape pairs       &  point pairs \\
    \hline
    \hline
    Airplane   & 500  & 9699  & 97.0M \\
    \hline
    Bag      &  76 & 1510  & 15.1M \\
    \hline
    Cap      &  55 & 1048  & 10.5M \\
    \hline
    Car      &  500 & 10000  & 100.0M \\
    \hline
    Chair      &  500 & 9997  & 100.0M \\
    \hline
    Earphone    &  69 & 1380  & 13.8M \\
    \hline
    Guitar    &  500 & 9962  & 99.6M \\
    \hline
    Knife    &  392 & 7821  & 78.2M \\
    \hline
    Lamp    &  500 & 9930  & 99.3M \\
    \hline
    Laptop    &  445 & 8880  & 88.8M \\
    \hline
    Motorbike  &  202 &  4040  & 40.4M \\
    \hline
    Mug    &  184 & 3680  & 36.8M \\
    \hline
    Pistol    &  275 & 5500  & 55.0M \\
    \hline
    Rocket    &  66 & 1320  & 13.2M \\
    \hline
    Skateboard    &  152 & 3032  & 30.3M \\
    \hline
    Table    &  500 & 9952  & 99.5M \\
    \hline
    \hline
    \end{tabular}%
  \caption{Our training dataset statistics.}
  \label{tab:train_stats}
\end{table}%


\paragraph{Network training.} All the parameters $\mathbf{w}$ of our deep architecture are estimated by minimizing a cost function, known as contrastive loss \cite{Hadsell:2006:DRL} in the literature of metric and deep learning. The cost function penalizes large descriptor differences for pairs of corresponding points, and small descriptor differences for pairs of non-corresponding points. We also include a regularization term in the cost function to prevent the parameter values from becoming arbitrarily large.  The cost function is formulated as follows:
\begin{equation}
\!\!L(\mathbf{w})\!=\!\!\!\sum\limits_{a,b \in C} \! D^2(X_a, X_b) +\! \sum\limits_{a,c \notin C} \!\! \max(m- D_{}(X_a, X_c),0)^2\
   + \! \lambda ||\mathbf{w}||^2
\end{equation}
where $C$ is a set of corresponding pairs of points derived from our part-based registration process and $D$ measures the Euclidean distance between a pair of input descriptors. The regularization parameter (known also as weight decay) $\lambda_1$ is set to $0.0005$. The quantity $m$, known as margin, is set to $1$ - its absolute value does not affect the learned parameters, but only scales distances such that non-corresponding point pairs tend to have a margin of at least one unit distance.

We initialize the parameters of the convolution layers (i.e., convolution filters) from  AlexNet~\cite{krizhevsky2012imagenet} trained on the ImageNet1K dataset (1.2M images)~\cite{ILSVRC15}. Since images contain shapes along with texture information, we expect that filters trained on massive image datasets already partially capture shape information. Initializing the network with filters pre-trained on image datasets proved successful in other shape processing tasks, such as shape classification \cite{SuMKL15}.

The cost function is minimized through batch gradient descent. At each iteration, $32$ pairs of corresponding points ${a,b \in C}$ are randomly selected. The pairs originate from random pairs of shapes for which our part-based registration has been executed beforehand. In addition, $32$ pairs of non-corresponding points ${a,c \notin C}$ are selected, making our total batch size equal to $64$. To update the parameters at each iteration, we use the Adam update rule~\cite{Adam}, which tends to provide faster convergence compared to  other  stochastic gradient descent schemes.

\paragraph{Implementation.} \rev{Our method is implemented using the Caffe deep learning library \cite{jia2014caffe}. Our source code, results and datasets are publically available on the project page:\\
 {\color{blue}\url{http://people.cs.umass.edu/~hbhuang/local_mvcnn/}}}.

\begin{figure*}[h]
    \centering
       \includegraphics[width=\textwidth]{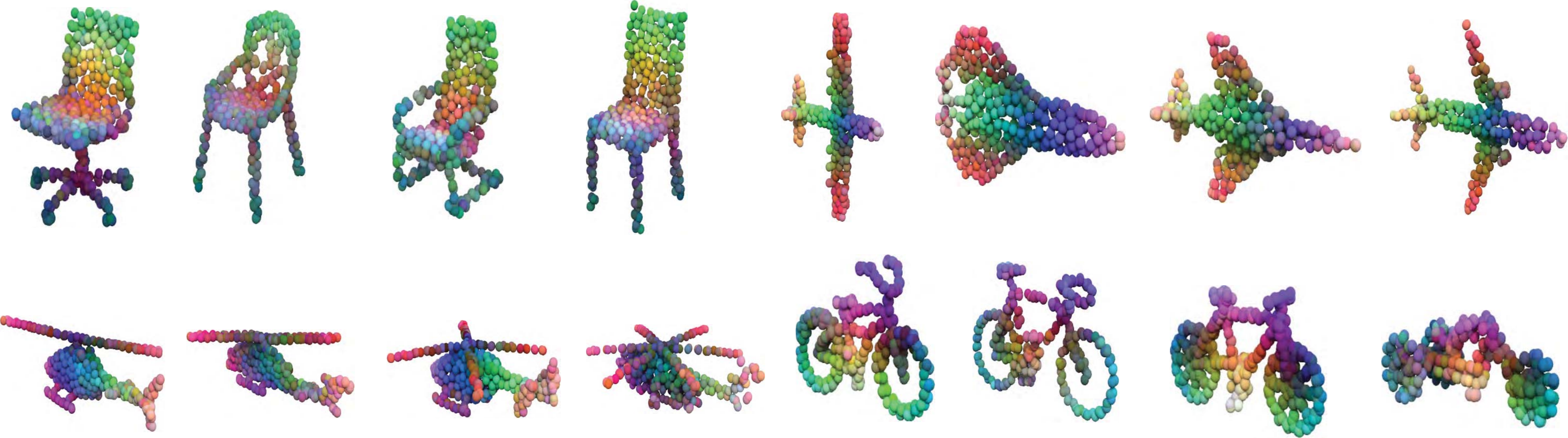}
       \caption{Visualization of point correspondence based on our learned descriptors for representative point-sampled BHCP shapes. Corresponding points have the same RGB\ color.}
    \label{fig:visualization_point}
\end{figure*}

\begin{figure*}[h]
    \centering
       \includegraphics[width=\textwidth]{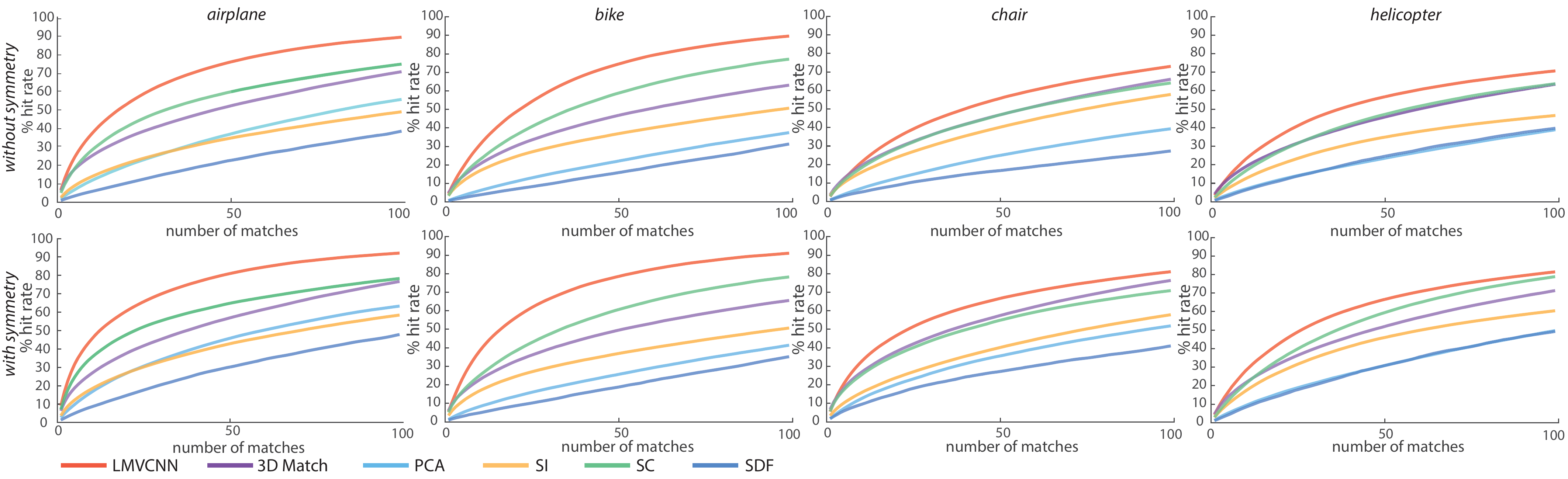}
       \caption{CMC plots for each category in the BHCP benchmark for all competing methods (single-category training for learning methods).}
    \label{fig:cmc_BHCP}
\end{figure*}
%
%
%
%
%
\begin{figure*}[h]
    \centering
       \includegraphics[width=\textwidth]{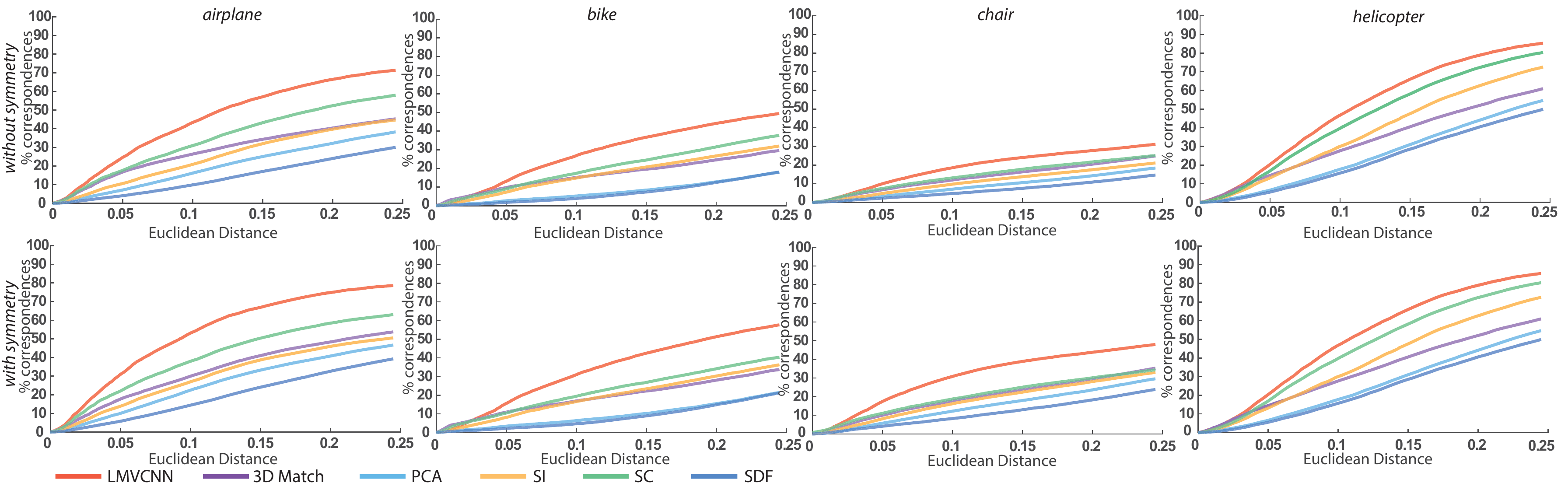}
       \caption{Correspondence accuracy for each category in the BHCP benchmark for all competing methods (single-category training for learning methods).}
    \label{fig:accuracy_BHCP}
\end{figure*}

\section{Evaluation}
\label{sec:evl}
\label{sec:results}
In this section we evaluate the quality of our learned local descriptors and compare them to state-of-the-art alternatives.

\paragraph{Dataset.}
We evaluate our descriptors on Kim et al.'s benchmark~\cite{kim2013learning}, known as the BHCP benchmark. The benchmark consists of  404 man-made shapes  including bikes,  helicopters, chairs, and airplanes originating from the Trimble Warehouse. The shapes have significant structural and geometric diversity.  Each shape has 6-12 consistently selected feature points with semantic correspondence (e.g. wingtips). Robust methods should provide descriptor values that discriminate these feature points from the rest, and embed corresponding points closely in descriptor space.

Another desired descriptor property is rotational invariance. Most shapes in BHCP are consistently upright oriented, which might bias or favor some descriptors. In general, 3D models available on the web, or in private collections, are not expected to always have consistent upright orientation, while existing algorithms to compute such orientation are not perfect even in small datasets (e.g. \cite{Fu08}). Alternatively, one could attempt to consistently align all shapes through a state-of-the-art registration algorithms \cite{Huang:2013:FSL}, however,  such methods often require human expert supervision or crowd-sourced corrections for large datasets \cite{Chang:2015}. To ensure that competing descriptors do not take advantage of any hand-specified  orientation or alignment in the data, and to test their rotational invariance, we apply a random 3D rotation to each BHCP shape. \rev{We also discuss results for our method when consistent upright-orientation is assumed (Section \ref{sec:different_algorithmic_choices}).}

\begin{figure*}[t!]
    \centering
                \includegraphics[width=1.0\textwidth]{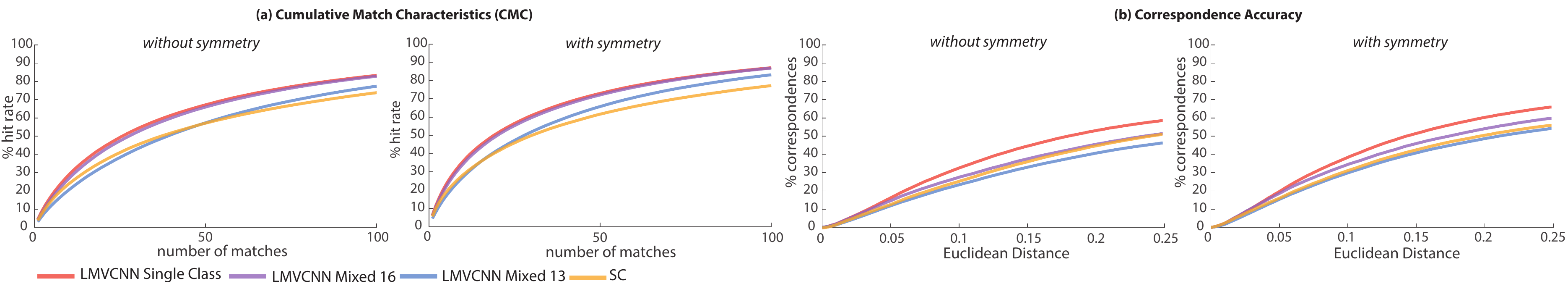}
   \caption{Correspondence accuracy for each category in the BHCP benchmark under cross-category training.}
    \label{fig:cross_category}
\end{figure*}

\paragraph{Methods.} \rev{We test our method against  various state-of-the-art techniques, including the learned descriptors  produced by the volumetric CNN of 3DMatch \cite{zeng20163dmatch}, and several hand-crafted alternatives: PCA-based descriptors used in \cite{kalogerakis2010learning,kim2013learning},  Shape Diameter Function (SDF) \cite{Shapira:2010:CPA}, geodesic shape contexts (SC) \cite{Belongie02,kalogerakis2010learning}, and Spin Images (SI) \cite{johnson1999using}. Although 3DMatch was designed for RGB-D images, the method projects the depth images back to 3D space to get Truncated Distance Function (TDF) values in a volumetric grid, where the volumetric CNN of that method operates on. We used the same type of input  for the volumetric CNN in our comparisons, by extracting voxel TDF patches around  3D surface points.  To ensure a fair comparison between our  approach and 3DMatch, we trained the volumetric CNN\ of the 3DMatch on the same training datasets as our CNN. We experimented with two training strategies for 3DMatch: (a) training their volumetric CNN from scratch on our datasets, and (b) initializing the volumetric CNN with their publicly available model, then fine-tuning it on our datasets. The fine-tuning strategy worked better than training their  CNN from scratch, thus we report results under this strategy.}

\paragraph{Training settings.} \rev{We evaluated our method against alternatives in two training settings. In our first  training setting, which we call the ``single-category'' setting, we train our method on the point-wise correspondence data (described in Section \ref{sec:learning}) from a single category and test on shapes of the same or another category. In an attempt to build a more generic descriptor, we also trained our method in a ``cross-category'' setting, for which we train our method on training data  across several categories of the segmented ShapeNetCore dataset, and test on shapes of the same or other categories. We discuss results for the ``single-category'' setting in Section \ref{sec:single_category_training}, and results for the  ``cross-category'' setting in Section \ref{sec:cross_category_training}}.

\paragraph{Metrics.} We use two popular measures to evaluate feature descriptors produced by all methods. First, we use the
\textit{Cumulative Match Characteristic (CMC)} measure which is designed to capture the proximity of corresponding points in descriptor space. In particular, given a pair of shapes, and an input feature point on one of the shapes, we retrieve a ranked list of points on the other shape. The list is ranked according to the Euclidean distance between these retrieved points and the input point in descriptor space. By recording the rank for all feature points across all pairs of shapes, we create a plot whose Y-axis is the fraction of ground-truth corresponding points whose rank is equal or below the  rank marked on the X-axis. Robust methods should assign top ranks to ground-truth corresponding points.

Another popular measure is  correspondence accuracy, also popularized  as the Princeton's protocol~\cite{kim2013learning}. This metric is designed to capture the proximity of predicted corresponding points to ground-truth ones in 3D space. Specifically, given a pair of shapes, and an input feature point on one of the shapes, we find the nearest feature point in descriptor space on the other shape, then measure the Euclidean distance between its 3D position and the position of the ground-truth corresponding point. By gathering Euclidean distances across all pairs of shapes, we create a plot whose  Y-axis  demonstrates  the  fraction  of  correspondences predicted correctly below a given Euclidean error threshold shown  on  the  X-axis. Depending on the application, matching symmetric points can be acceptable. Thus, for both metrics, we discuss below results where we accept symmetric (e.g. left-to-right wingtip) matches, or not accepting them.

\begin{figure*}[t]
    \centering
       \includegraphics[width=1.0\textwidth]{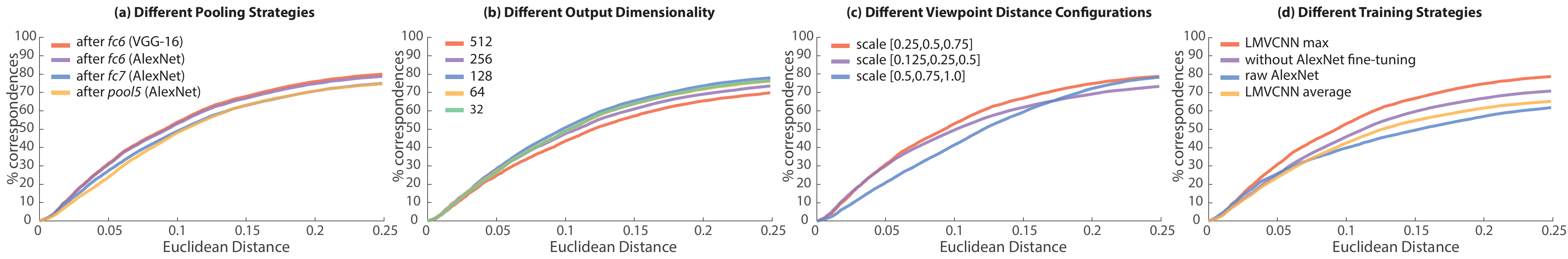}
       \caption{Evaluation of alternative algorithmic choices on the BHCP dataset (airplanes).}
    \label{fig:choices}
\end{figure*}

\begin{table*}[t]
\centering
\begin{tabular}{|@{}c@{}|@{}c@{}|@{}c@{}|@{}c@{}|@{}c@{}|@{}c@{}|@{}c@{}|@{}c@{}|@{}c@{}|}
\hline
Method & \,ours (single class)\, & \,ours (`Mixed 16')\, & \,ours (`Mixed 13')\, & \,3DMatch (single class)\, & \,\,\,PCA\,\,\, & \,\,\,\,\,SI\,\,\,\,\, & \,\,\,\,\,SC\,\,\,\,\, & \,\,\,\,SDF\,\,\, \\

\hline
CMC  & \multirow{2}{*}{87.1\%} & \multirow{2}{*}{86.9\%} & \multirow{2}{*}{83.2\%} & \multirow{2}{*}{66.2\%}   & \multirow{2}{*}{43.3\%} & \multirow{2}{*}{51.2\%} & \multirow{2}{*}{77.3\%} & \multirow{2}{*}{34.5\%} \\
(symmetry) & & & & & & & & \\
\hline

\hline
CMC  & \multirow{2}{*}{83.3\%} & \multirow{2}{*}{82.8\%} & \multirow{2}{*}{77.4\%} & \multirow{2}{*}{47.6\%}   & \multirow{2}{*}{39.8\%} & \multirow{2}{*}{49.8\%} & \multirow{2}{*}{73.8\%} & \multirow{2}{*}{35.3\%} \\
(no symmetry) & & & & & & & & \\

\hline
Corr. accuracy  & \multirow{2}{*}{65.9\%} & \multirow{2}{*}{59.8\%} & \multirow{2}{*}{54.1\%} & \multirow{2}{*}{47.6\%}   & \multirow{2}{*}{39.8\%} & \multirow{2}{*}{49.8\%} & \multirow{2}{*}{56.1\%} & \multirow{2}{*}{35.3\%} \\
(symmetry) & & & & & & & & \\
\hline

\hline
Corr. accuracy  & \multirow{2}{*}{58.5\%} & \multirow{2}{*}{51.3\%} & \multirow{2}{*}{46.2\%} & \multirow{2}{*}{40.6\%}   & \multirow{2}{*}{32.7\%} & \multirow{2}{*}{43.1\%} & \multirow{2}{*}{50.5\%} & \multirow{2}{*}{28.5\%} \\
(no symmetry) & & & & & & & & \\
\hline

\hline

\end{tabular}%
\caption{Numerical results for correspondence accuracy and CMC for all methods averaged over the BHCP dataset. We include the single-category and cross-category settings of our method. Correspondence accuracy is reported based on fraction of correspondences predicted correctly below Euclidean distance error threshold 0.25. CMC\ is reported for 100 retrieved matches. }
\label{tab:num_average_accuracy}
\end{table*}%

\subsection{Results: single-category training.}
\label{sec:single_category_training}
\rev{In this setting, to test our method and 3DMatch on BHCP\ airplanes, we train both methods on training correspondence data from ShapeNetCore airplanes. Similarly, to test on BHCP chairs, we train both methods on  ShapeNetCore chairs. To test on BHCP bikes, we train both methods on ShapeNetCore bikes. Since both the BHCP and ShapeNetCore shapes originate from 3D\ Warehouse, we ensured that the test BHCP shapes were excluded from our training datasets. There is no helicopter class in ShapeNetCore, thus to test on BHCP helicopters, we train both methods on ShapeNetCore airplanes, a related but different class. We believe that this test on helicopters is particularly interesting since it demonstrates the generalization ability of the learning methods to another class. We note that the hand-crafted descriptors are class-agnostic, and do not require any training, thus we simply evaluate them on the BHCP shapes.}

\rev{Figures \ref{fig:cmc_BHCP} demonstrates the CMC plots for all the methods on the BHCP dataset for both symmetric and non-symmetric cases (we refer to our method as `local MVCNN', or in short `LMVCNN'). Figure \ref{fig:accuracy_BHCP} shows the corresponding plots for the corresponding accuracy measure. \revB{Table \ref{tab:num_average_accuracy} reports the evaluation measures numerically for all methods.}
 According to both the  CMC and correspondence accuracy metrics, and in both symmetric and non-symmetric cases, we observe that our learned descriptors outperform the rest, including the learned descriptors of 3DMatch, and the hand-engineered local descriptors commonly used in 3D\ shape analysis. Based on these results, we believe that our method successfully embeds semantically similar feature points  in descriptor space closer than other methods. Figure \ref{fig:visualization_point}
visualizes predicted point correspondences produced by our method for the BHCP test shapes. We observed that our predicted correspondences appear visually plausible, although for bikes we also see some inconsistencies (e.g., at the pedals).  We believe that this happens because our automatic non-rigid alignment method tends to produce less accurate training correspondences for the parts of these shapes whose geometry and topology vary significantly. In the supplementary material, we include examples of correspondences computed  between BHCP shapes.}

\subsection{Results: cross-category training.}
\label{sec:cross_category_training}
\rev{In this setting, we train our method on the training correspondence data generated for all $16$ categories of the segmented ShapeNetCore dataset ($\sim$977M correspondences), and evaluate on the BHCP shapes (again, we ensured that the test BHCP shapes were excluded from this training dataset.) Figure~\ref{fig:cross_category}(a) demonstrates the CMC plots and Figure~\ref{fig:cross_category}(b) demonstrates the correspondence accuracy plots for our method trained across all 16 ShapeNetCore categories (``Mixed 16'') against the best performing alternative descriptor (shape contexts)  for the symmetric and non-symmetric case averaged over all BHCP\ classes (see supplementary material for plots per class). As a reference, we also include the plots for our method trained in the single-category setting (``Single Class'').
 We observed that the performance of our method slightly drops in the case of cross-category training, yet still significantly outperforms the best alternative method.}

\revB{We further stretched the evaluation of our method to the case where we train it on $13$ categories of the segmented ShapeNetCore (``Mixed 13''), excluding airplanes, bikes, and chairs, i.e. the categories that also exist in BHCP.  We observe that the performance of our method is still comparable to the SC descriptor (higher in terms of CMC, but a bit lower in terms of correspondence accuracy). This means that in the worst case where our method is tested on shape categories not  observed during training, it can  still produce fairly general local shape descriptors that perform favorably compared to hand-crafted alternatives.}

%

\subsection{Results: alternative algorithmic choices.}
\label{sec:different_algorithmic_choices}
\rev{In Figure~\ref{fig:choices}, we demonstrate correspondence accuracy (symmetric case) under different choices of architectures and viewpoint configurations for our method. Specifically, Figure \ref{fig:choices}(a) shows results with view-pooling applied after the pool5, fc6 and fc7 layer of AlexNet. View-pooling after fc6 yields the best performance. We also demonstrate results with an alternative deeper network, known as VGG16 \cite{Simonyan14c}, by applying view-pooling after its fc6 layer. Using VGG16 instead of AlexNet offers marginally better performance than AlexNet, at the expense of slower training and testing. Figure \ref{fig:choices}(b) shows results with different output dimensionalities for our descriptor, and Figure \ref{fig:choices}(c) shows results with different viewpoint distance configurations. Our proposed configuration offers the best performance. Figure \ref{fig:choices}(d) shows results when we fix the AlexNet layers and update only the weights for our dimensionality reduction layer during training (``no AlexNet fine-tuning''), and when we remove the dimensionality reduction layer and we just perform view-pooling on the raw 4096-D features produced by AlexNet again without fine-tuning (``raw AlexNet''). It is evident that fine-tuning the AlexNet layers and using the dimensionality reduction layer are both useful to achieve high performance.  Figure \ref{fig:choices}(d)  shows performance when ``average'' view pooling is used in our architecture instead of ``max'' view pooling. ``Max'' view pooling offers significantly higher performance than ``average'' view pooling.}

%
\begin{figure}[h!]%
  \includegraphics[width=0.6\linewidth]{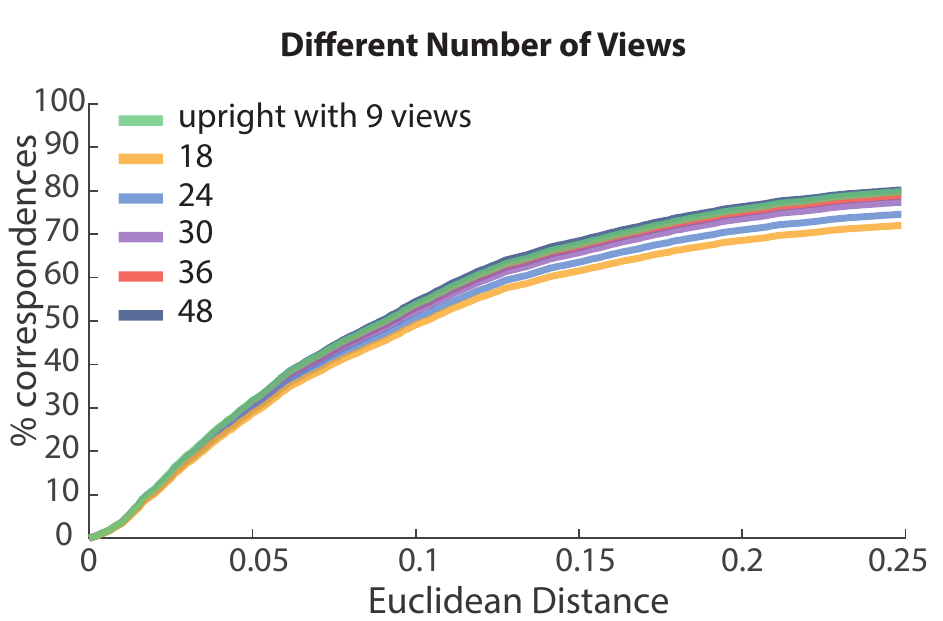}
  \caption{Evaluation wrt various number of views, and upright orientation on the BHCP dataset (airplanes)}
  \label{fig:choices_view}  
\end{figure}

Figure \ref{fig:choices_view} demonstrates results under different numbers of sampled views, and results assuming consistent upright orientation (i.e. we apply random rotation to BHCP\ shapes only about the upright axis). The plots indicate that the performance of our method is very similar for both the arbitrary orientation and consistent upright orientation cases (slightly higher in the upright orientation case, which makes sense since the test views would tend to be more similar to the training ones in this case). In the presence of fewer sampled views (e.g. $18$ or $24$), the performance of our method drops slightly, which is an expected behavior since less surface information is captured. The performance of our method is quite stable beyond $30$ views. Table \ref{tab:test_time} reports the execution times to compute our descriptor per point with respect to various number of views. The reported execution times include both the stage for rendering views and the stage for processing these renderings through our network. We observe that the execution time tends to scale linearly with the number of views. We also note that multiple surface points could be processed in parallel.

\begin{table}[h!]
\centering
\begin{tabular}{|c|c|c|c|c|c|}
\hline
view size & 18 & 24 & 30 & 36 & 48\\
\hline
execution time & 0.27s  & 0.44s  & 0.51s  & 0.57s & 0.91s\\
\hline
\end{tabular}%
\caption{Execution time  wrt different  numbers of used views}
\label{tab:test_time}
\end{table}%


\begin{figure}[b]
    \centering
    \includegraphics[width=\linewidth]{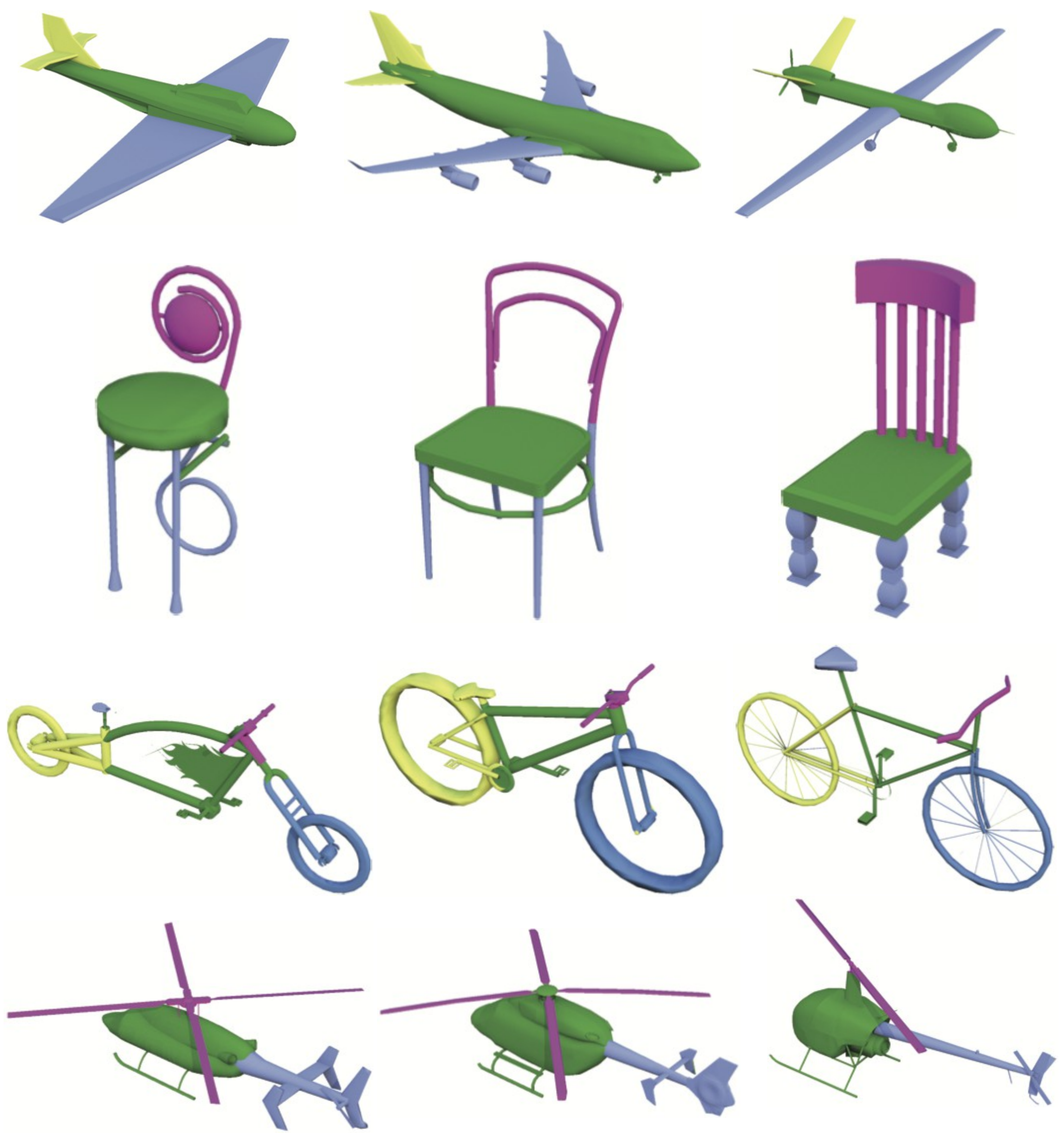}
    \caption{Examples of shape segmentation results on the BHCP dataset.}
    \label{fig:segmentation}
\end{figure}

\begin{figure*}[t]
    \centering
    \includegraphics[width=\textwidth]{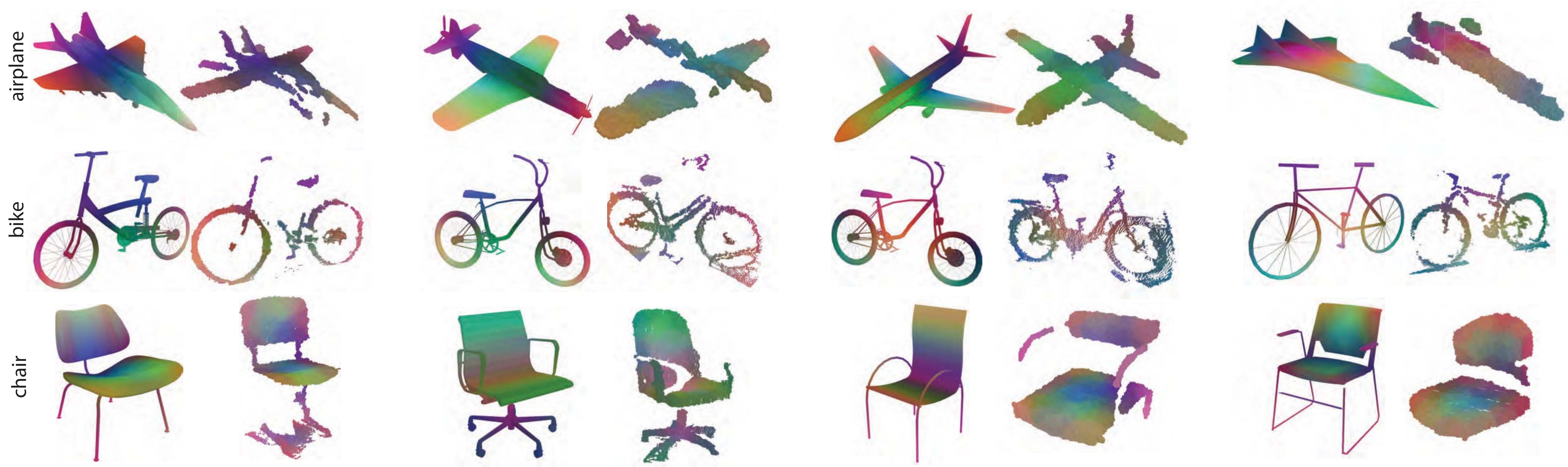}
    \caption{Dense matching of partial, noisy scans (even columns) with 3D complete database shapes (odd columns). Corresponding points have consistent colors.}
    \label{fig:scan_match}
\end{figure*}

\section{Applications}
\label{sec:app}

In this section, we utilize our learned point descriptors for a wide variety of shape analysis applications, and evaluate with respect to existing methods, and benchmarks. Specifically, we discuss applications of our descriptors to shape segmentation, affordance region detection, and finally matching depth data with 3D models.

\begin{table}[t]
\centering
\begin{tabular}{@{}c@{}|@{}c@{}|@{}c@{}|@{}c@{}}
\multirow{2}{*}{Category} & \multirow{1}{*}{JointBoost} & \multirow{1}{*}{JointBoost} & \multirow{2}{*}{Guo et al.}\\
    \,                      & \,our descriptors\,                      &  \,hand-crafted\,       &  \,\\
    \hline
    \hline
    Bikes        & \textbf{77.3\%} &  72.4\%  & 69.6\% \\
    \hline
    Chairs      &  \textbf{71.8\%}  & 65.7\%  & 60.1\% \\
    \hline
    Helicopters &  91.7\%          & 91.1\%  & \textbf{95.6\%} \\
    \hline
    Airplanes   & \textbf{85.8\%}  & 85.1\%  & 81.7\% \\
    \hline
    \hline
    Average     & \textbf{81.7\%}  & 78.6\%  & 76.7\% \\
    \end{tabular}%
  \caption{Mesh labeling accuracy on BHCP test shapes.}
  \label{tab:segmentation}
\end{table}%

\paragraph{Shape segmentation.}
We first demonstrate how our descriptors can benefit shape segmentation. \rev{Given an input shape, our goal is to use our descriptors to label surface points according to a set of part labels}. We follow the graph cuts energy formulation by \cite{kalogerakis2010learning}. The graph cuts energy relies on unary terms that assesses the consistency of mesh faces with part labels, and pairwise terms that provide cues to whether adjacent faces should have the same label. To evaluate the unary term, the original implementation relies on local hand-crafted descriptors computed per mesh face. The descriptors include surface curvature, PCA-based descriptors, local shape diameter, average geodesic distances, distances from medial surfaces, geodesic shape contexts, and spin images. We replaced all these hand-crafted descriptors with descriptors extracted by our method  to check whether segmentation results are improved.

Specifically, we trained our method on ShapeNetCore classes as described in the previous section, then extracted descriptors for $256$ uniformly sampled surface points for each shape in the corresponding test classes of the BHCP dataset. Then we trained a JointBoost classifier using the  same hand-crafted descriptors used in \cite{kalogerakis2010learning} and our descriptors. We also trained the CNN-based classifier proposed in \cite{Guo:2015}.  This method proposes to regroup the above hand-crafted descriptors in a $30$x$20$ image, which is then fed into a CNN-based classifier. Both classifiers were trained on the same training and test split. We used 50\% of the BHCP\ shapes for training, and  the other 50\% for testing per each class. The classifiers extract per-point probabilities, which are then projected back to nearest mesh faces to form the unary terms used in graph cuts.

We measured labeling accuracy on test meshes for all methods (JointBoost with our learned descriptors and graph cuts, JointBoost with hand-crafted descriptors and graph cuts, CNN-based classifier on hand-crafted descriptors with graph cuts). Table \ref{tab:segmentation} summarizes the results. Labeling accuracy is improved on average with our learned descriptors, with significant gains for chairs and bikes in particular.




\paragraph{Matching shapes with 3D\ scans.}
\label{sec:match}
Another application of our descriptors is dense matching between scans and 3D models, which can in turn benefit shape and scene understanding techniques.  Figure \ref{fig:scan_match} demonstrates dense matching of partial, noisy scanned shapes with manually picked 3D\ database shapes for a few characteristic cases. Corresponding (and symmetric) points are visualized with same color. \rev{Here we trained our method on ShapeNetCore classes in the single-category training setting, and extracted descriptors for input scans and shapes picked from the BHCP\ dataset. Note that we did not fine-tune our network on scans or point clouds. To render point clouds, we use a small ball centered at each point. Even if the scans are noisy, contain outliers,  have entire parts missing, or have noisy normals and consequent shading artifacts, we found that our method can still produce robust descriptors to densely match them with complete shapes.}

\begin{figure}[h]
    \centering
    \includegraphics[width=\linewidth]{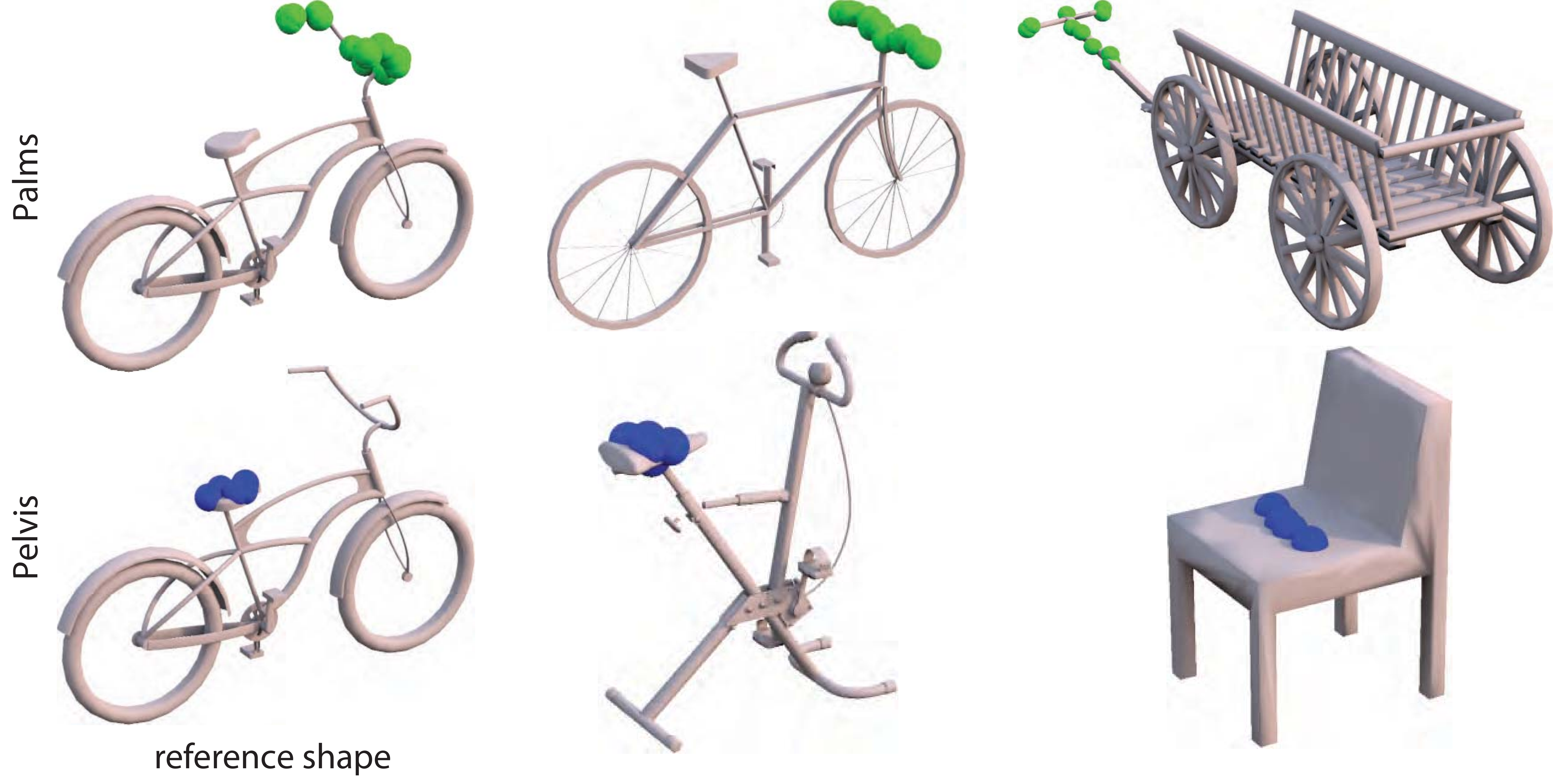}
    \caption{Corresponding affordance regions for pelvis and palms.}
    \label{fig:contact_affordances}
\end{figure}

\paragraph{Predicting affordance regions.}
Finally, we demonstrate how our method can be applied to predict human affordance regions on 3D shapes. Predicting affordance regions is particularly challenging since regions across shapes of different functionality should be matched (e.g. contact areas for hands on a shopping cart, bikes, or armchairs). To train and evaluate our method, we use the affordance benchmark with manually selected contact regions for people interacting with various objects~\cite{Kim14s2p} (e.g. contact points for pelvis and palms). \rev{Starting from our model trained in the cross-category setting}, we fine-tune it based on corresponding regions marked in a training split we selected from the benchmark (we use $50\%$ of its shapes for fine-tuning). The training shapes are scattered across various categories, including bikes, chairs, carts, and gym equipment. Then we evaluate our method by extracting descriptors for the rest of the shapes on the benchmark.
Figure \ref{fig:contact_affordances} visualizes corresponding affordance regions for a few shapes for pelvis and palms. Specifically, given marked points for these areas on a reference shape (first column), we retrieve points on other shapes based on their distance to the marked points in our descriptor space. As we can see from these results, our method can also generalize to matching local regions across shapes from different categories with largely different global structure. We refer the reader to the supplementary material for more results.
%
%

\section{Conclusion}

We presented a method that computes local shape descriptors by taking multiple rendered views of  shape regions in multiple scales and processing them through a learned deep convolutional network. Through view pooling and dimensionality reduction, we produce compact local descriptors that can be used in a variety of shape analysis applications. Our results confirm the benefits of using such view-based architecture. We also presented a strategy to  generate  synthetic training data to automate the learning procedure.

\begin{figure}[b!]
    \centering
    \includegraphics[width=1.0\linewidth]{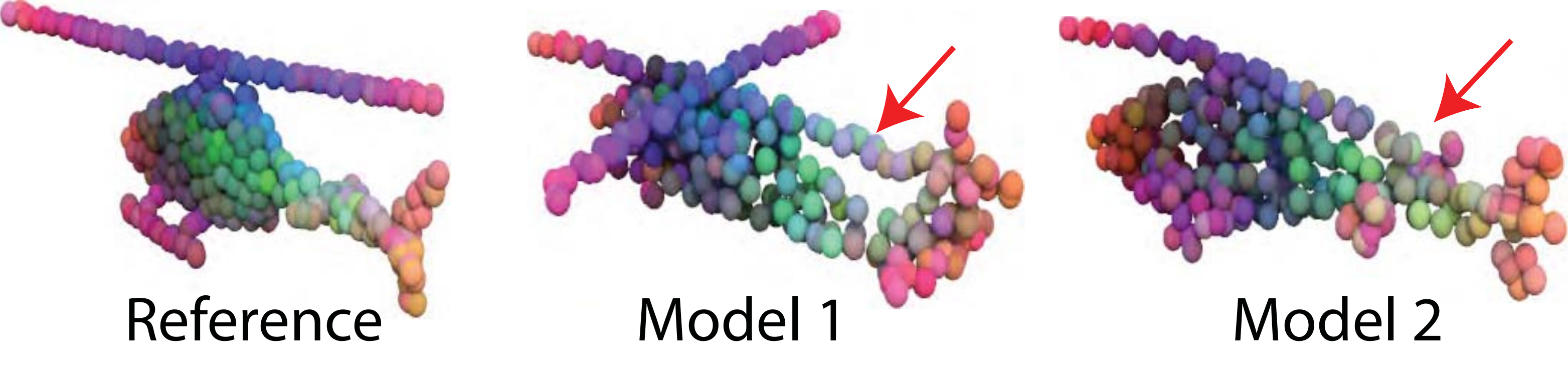}
    \caption{Our learned descriptors   are less effective in shape classes for which training correspondences tend to be erroneous.    }
    \label{fig:limitation}
\end{figure}

There are a number of avenues of future directions that can address limitations of our method. Currently, we rely on a heuristic-based viewing configuration and rendering procedure. It would be interesting to investigate optimization strategies to automatically select best viewing configurations and rendering styles to maximize performance. We currently rely on  perspective projections to capture local surface information. Other local surface parameterization schemes  might be able to capture more surface information that could be further processed through a deep network. \rev{Our automatic non-rigid alignment method tends to produce less accurate training correspondences for parts of training shapes whose geometry and topology vary significantly. Too many erroneous training correspondences will in turn affect the discriminative performance of our descriptors (Figure \ref{fig:limitation}). Instead of relying on synthetic training data exclusively, it would be interesting to explore crowdsourcing techniques for gathering human-annotated correspondences in an active learning setting.} Rigid or non-rigid alignment methods could benefit from our descriptors, which could in turn improve the quality of the training data used for learning our architecture. This indicates that iterating between training data generation, learning, and non-rigid alignment could further improve performance. \revB{Finally,  zero-shot learning techniques \cite{XianSA17} also represent an interesting avenue to improve the generalization of data-driven surface descriptors.}

\section{Acknowledgements}
\rev{Kalogerakis acknowledges support from NSF (CHS-1422441, CHS-1617333), NVidia and Adobe. Chaudhuri acknowledges support from Adobe and Qualcomm. Our experiments were performed in the UMass GPU cluster obtained under a grant from the Collaborative R\&D Fund managed by the Massachusetts Technology Collaborative}

\bibliographystyle{ACM-Reference-Format}
\bibliography{point_feature}
\end{document}